\documentclass{ceurart}
\usepackage{listings}
\usepackage{float}
\usepackage{placeins}

\begin{document}

\copyrightyear{2025}
\copyrightclause{Copyright for this paper by its authors.
  Use permitted under Creative Commons License Attribution 4.0
  International (CC BY 4.0).}

\conference{CLEF 2026: Conference and Labs of the Evaluation Forum, September 21-24, 2026, Jena, Germany}

\title{Retrieval-Augmented Biomedical Question Answering with Weak-Question Recovery and Neural Reranking for BioASQ Task 14b}

\author[1]{Xueying Zhao}[
    orcid=0000-0003-4184-7040,
    email=xzhao348@gatech.edu,
]
\cormark[1]

\author[1]{Lee Mai}[
    email=lmai39@gatech.edu,
]
\author[1]{Balaji Anandganesh}[
    email=banandganesh3@gatech.edu,
]

\address[1]{Georgia Institute of Technology, North Ave NW, Atlanta, GA 30332}
\cortext[1]{Corresponding author.}

\begin{abstract}
    This work presents DS@GT ARC BioASQ team's work for a biomedical question answering pipeline, integrating multi-source query expansion, neural reranking, retrieval refinement, and OpenBioLLM–assisted answer generation. The system combines PubMed retrieval with fine-tuned MiniLM-based semantic reranking, Reciprocal Rank Fusion (RRF), and feature-based relevance scoring to improve document ranking quality. To address challenging queries with weak retrieval performance, we introduce a conditional weak-question recovery strategy that applies semantic expansion, relationship-aware augmentation, and selective result merging. A post-retrieval pruning stage further removes redundant or low-relevance snippets while preserving evidence coverage for downstream answer generation. Experimental results on BioASQ evaluation batches demonstrate that the proposed recovery and cleanup strategies substantially improve retrieval robustness and MAP@10 performance on difficult question sets. The final system also incorporates output validation and post-processing steps to ensure formatting consistency and submission reliability across BioASQ phases.
\end{abstract}

\begin{keywords}
  BioASQ \sep
  biomedical question answering \sep
  information retrieval \sep
  large language models \sep 
  OpenBioLLM \sep
  retrieval-augmented generation
\end{keywords}

\maketitle

% We recommend splitting your main document into smaller parts that are easier to navigate.
% The input command "imports" the contents of the file into the current location.
% Prefixing the document with a number allows for natural string sorts.
\section{Introduction}

Biomedical question answering (QA) aims to automatically retrieve and synthesize reliable evidence from large-scale biomedical literature in response to natural language questions. As the volume of biomedical publications continues to grow rapidly, manually identifying relevant evidence from resources, such as PubMed, becomes increasingly difficult for researchers and clinicians. Benchmark challenges such as BioASQ have therefore become important platforms for evaluating systems capable of retrieving relevant biomedical documents and generating accurate evidence-based answers \cite{tsatsaronis2015,nentidis2020bioasq}

BioASQ Task 14b is divided into multiple phases with increasing complexity. Phase A focuses on
document and snippet retrieval, while Phase A+ and Phase B additionally require systems to generate exact answers and ideal summary-style responses. Although recent advances in large language models (LLMs) have significantly improved text generation quality, retrieval quality remains a major bottleneck in biomedical QA. Retrieved documents may contain partial, noisy, or weakly related evidence, and
high retrieval relevance does not always translate into correct downstream answers.

Beyond benchmark performance, biomedical question answering systems also play an increasingly important role in improving scientific accessibility, translational healthcare research, and evidence-driven biomedical workflows. Recent advances in biomedical artificial intelligence have explored intelligent automation, machine learning-assisted healthcare analysis, and AI-integrated biomedical systems across domains including multiomics, healthcare inequity assessment, and scientific communication support \cite{zhao2026breastcancer,zhao2026microfluidics,zhao2026multiomics,hayslett2024sciencecommunication}. These developments further emphasize the importance of robust biomedical retrieval systems capable of supporting reliable evidence synthesis and knowledge discovery from rapidly growing biomedical literature.

Recent BioASQ systems increasingly adopt retrieval-augmented generation (RAG) frameworks that combine sparse retrieval, dense retrieval, neural reranking, and LLM-based reasoning \cite{stuhlmann2025efficient,hu2024serts}. More recent studies further explore agentic retrieval strategies and dynamic biomedical knowledge graphs to improve reasoning over complex biomedical relationships \cite{rezaei2025agentic}. Self-reflective retrieval refinement and feedback-driven retrieval loops have also been investigated to reduce hallucinations and improve evidence grounding \cite{ateia2025selffeedback}. However, biomedical QA remains challenging due to domain-specific terminology, synonym variability, entity ambiguity, and multi-hop reasoning requirements. Furthermore, LLM-generated outputs may become inconsistent when retrieval evidence is weak or incomplete. These challenges motivate the need for more robust retrieval and evidence-selection pipelines.

In this work, we develop a biomedical QA pipeline that integrates multi-source query expansion, retrieval fusion, MiniLM-based neural reranking, weak-question recovery, and post-retrieval pruning strategies. Instead of relying solely on a single retrieval formulation, our system combines training-memory expansion, biomedical synonym enrichment using OpenBioLLM \cite{OpenBioLLMs}, and relationship-aware augmentation to improve recall across diverse biomedical question types. We further introduce a conditional weak-question recovery mechanism that selectively revisits poorly performing queries using semantic and answer-guided refinement strategies.
\section{Related Work}

Early biomedical QA systems primarily relied on keyword-based information retrieval approaches, such as BM25 combined with heuristic ranking and rule-based answer extraction. These approaches were computationally efficient and interpretable, but often struggled with synonym variation, semantic ambiguity, and complex biomedical terminology.

More recent systems incorporate dense retrieval and neural reranking techniques to improve semantic matching between questions and biomedical documents. Transformer-based encoders, such as Bioformer \cite{fang2023bioformer} and sequence-to-sequence reranking approaches \cite{nogueira2020seq2seq,reimers2019sentencebert}, have been widely adopted for retrieval and reranking tasks due to their ability to capture contextual semantic similarity beyond lexical overlap. Hybrid retrieval frameworks that combine sparse retrieval with dense semantic reranking have become increasingly common in biomedical QA and RAG systems \cite{stuhlmann2025efficient}.Biomedical QA benchmarks and curated datasets such as BioASQ-QA have further accelerated the development of retrieval-augmented biomedical question answering systems by providing large-scale expert-annotated biomedical question-answer pairs and evidence documents \cite{pappas2022bioasq}.

The rapid development of LLMs has further expanded the capabilities of biomedical QA systems. RAG approaches combine retrieved evidence with LLM-based answer synthesis, enabling systems to generate more coherent and context-aware responses. Recent work has explored agentic retrieval frameworks, dynamic knowledge graph construction, ontology-grounded retrieval, and self-reflective retrieval loops to improve reasoning over biomedical evidence \cite{rezaei2025agentic,ateia2025selffeedback,sharma2025ograg}. These methods aim to reduce hallucinations and improve factual grounding by iteratively refining retrieval results before answer generation. Tree-search and self-rewarding retrieval strategies have also been proposed to improve reasoning quality in biomedical RAG systems \cite{hu2024serts}. In parallel, orchestration frameworks, such as LangGraph have enabled more flexible agentic workflows involving iterative planning, retrieval, and feedback cycles \cite{sapkota_agentic}.

Despite these advances, several practical challenges remain. Biomedical retrieval pipelines frequently produce noisy or weakly relevant snippets, especially for rare diseases, complex relationships, or multi-hop questions. Additionally, LLM-generated outputs may become inconsistent when evidence quality is insufficient. Many systems therefore require additional post-processing, filtering, or evidence-selection mechanisms to ensure stable downstream answer generation.

Our work builds upon these prior retrieval-augmented approaches while emphasizing retrieval robustness and recovery mechanisms. Rather than relying solely on a fixed retrieval pipeline, we incorporate multi-source query expansion, MiniLM-based reranking, weak-question recovery, and safe pruning strategies to improve evidence quality before answer generation.

\section{BioASQ Task 14b Dataset}
We conducted our experiments using the BioASQ Task 14b (CLEF 2026) biomedical question answering dataset. The benchmark supports retrieval evaluation (Phase A) and answer generation (Phases A+ and B), covering four question types: yes/no, factoid, list, and summary. The training set contains 5,729 annotated questions, each paired with PubMed documents, evidence snippets, ideal answers, and type-specific exact answers. These annotations were used for query expansion, reranker training, retrieval analysis, and answer generation.

Official evaluation was performed on the BioASQ test batches. During the competition, Phase A evaluates document retrieval, while Phases A+ and B evaluate answer generation using the official supporting documents and snippets provided by the organizers, without access to reference answers.

\section{Methodology}
\subsection{Retrieval and Reranking Pipeline}
Figure 1 illustrates the overall retrieval and reranking framework used in our system. The retrieval pipeline begins by expanding each question using domain-specific strategies. These include training-memory expansion based on similar BioASQ questions, biomedical synonym expansion using OpenBioLLM, and relationship-aware augmentation that capture common associations such as disease–treatment or gene–pathway links. These complementary expansions generate a diverse set of enriched queries.

\begin{figure}[htbp]
    \centering
    \includegraphics[width=1\linewidth]{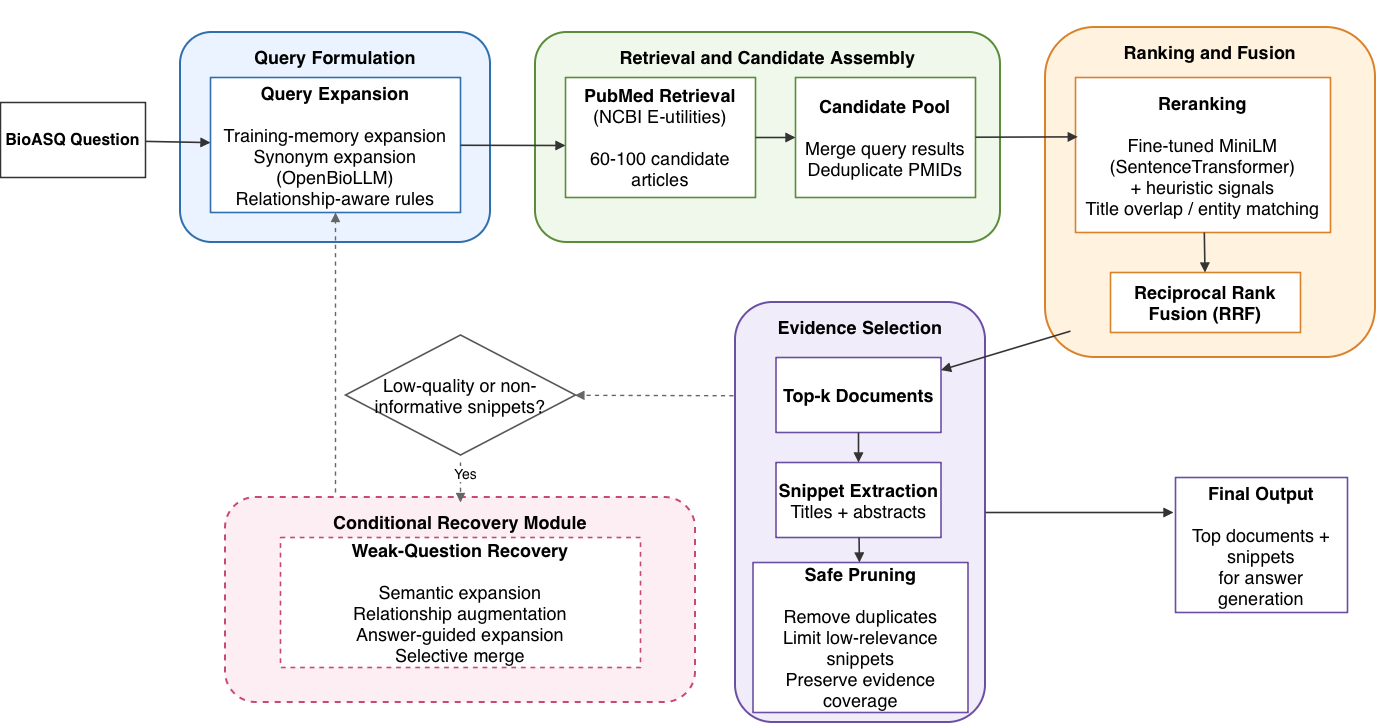}
    \caption{Overview of the proposed BioASQ retrieval pipeline. The system combines multi-source query expansion, PubMed retrieval, fine-tuned MiniLM reranking, Reciprocal Rank Fusion (RRF), snippet extraction, and safe pruning. A conditional weak-question recovery module further improves difficult queries using semantic and relationship-aware expansion strategies. Solid arrows indicate the primary retrieval flow, while dashed arrows represent the conditional weak-question recovery process.
}
    \label{fig:placeholder}
\end{figure}

The expanded queries are submitted to the PubMed API (via NCBI E-utilities) to retrieve candidate documents, typically in the range of 60 to 100 articles per question. Retrieved results from multiple queries are merged into a unified candidate pool, with duplicate PMIDs removed to ensure diversity. The candidate documents are then reranked using a fine-tuned MiniLM-based SentenceTransformer model, which captures semantic similarity between the query and document content. In addition to neural similarity scoring, heuristic signals, such as title overlap, key phrase matching, and entity-level alignment, are incorporated to further refine ranking. Results from multiple query expansions are combined using Reciprocal Rank Fusion (RRF), improving robustness across query variations.

To address cases where the initial retrieval fails to return sufficiently relevant documents, we introduce a weak-question recovery step. Specifically, we identify a subset of poorly performing queries based on snippet quality and retrieval signals. The retrieval signals include the number of unique retrieved candidate articles, semantic relevance after MiniLM reranking, snippet redundancy, and lexical overlap between the question and the retrieved evidence. During system development, snippet quality was assessed using a composite lexical relevance score that combines keyword recall, entity overlap, phrase matching, and topic-drift penalties. Empirically, snippets with scores below 0.35 were considered weak evidence, whereas scores above 0.55 were treated as strong evidence. Questions dominated by weak evidence or exhibiting limited retrieval coverage were therefore selected for recovery. For these cases, we apply targeted semantic expansion, relationship-aware augmentation, and domain-specific synonym enrichment. Rather than replacing the entire retrieval output, improved results are selectively merged when they demonstrate clear gains in relevance. Finally, a post-retrieval pruning strategy is applied to remove irrelevant or redundant snippets while preserving sufficient evidence coverage. This step helps balance precision and recall, ensuring that the final top-ranked documents and extracted snippets are both informative and concise for downstream answer generation.

\subsection{Snippets Selection and Pruning}
To better understand the relationship between snippet count and evidence coverage, we analyzed token overlap between retrieved snippets and BioASQ ideal answers across the training set. Figure 2 shows the average vocabulary coverage as the number of snippets increases. While coverage continues to improve with additional snippets, the gains begin to diminish after approximately two snippets per question. This observation motivated the use of controlled snippet pruning and selective evidence filtering in the final pipeline. Using fewer high-quality snippets also reduces retrieval noise and improves prompt efficiency during downstream answer generation. Based on this analysis, we prioritize concise evidence selection rather than maximizing the total number of retrieved snippets.

\begin{figure}[htbp]
    \centering
    \includegraphics[width=1\linewidth]{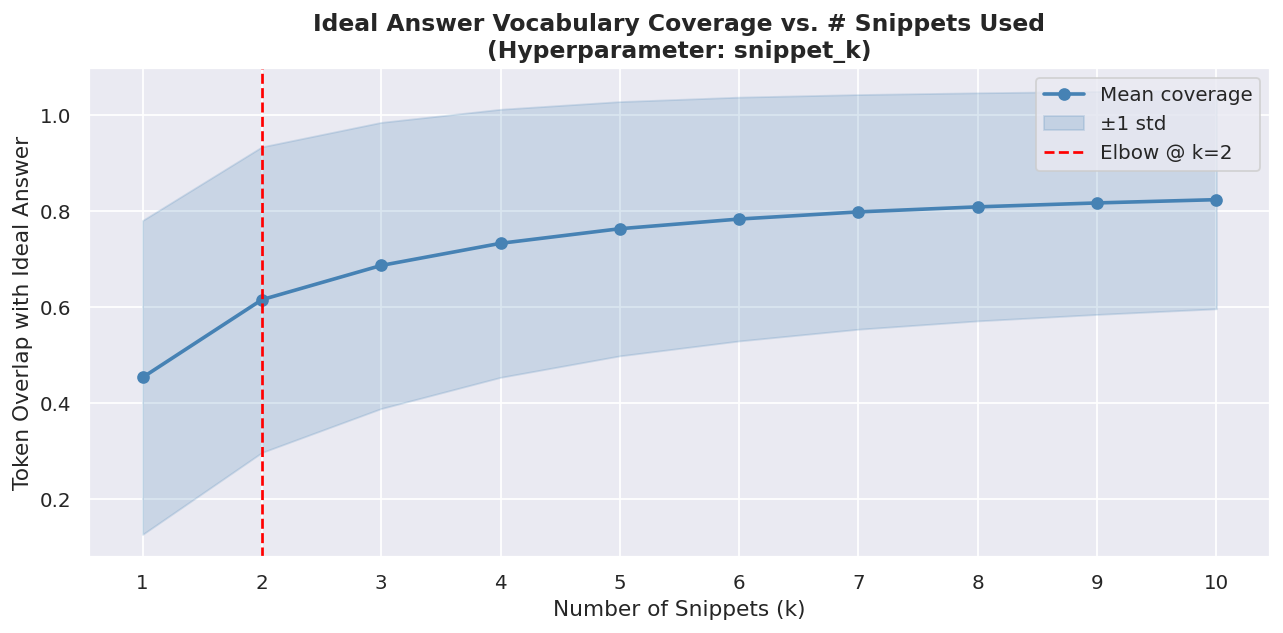}
    \caption{Relationship between the number of retrieved snippets and ideal-answer vocabulary coverage across the BioASQ training set. Coverage improves rapidly for the first few snippets but shows diminishing returns beyond approximately two snippets, motivating controlled snippet pruning and selective evidence filtering in the final retrieval pipeline. The shaded region represents one standard deviation across questions.}
    \label{fig:placeholder}
\end{figure}
\subsection{Neural Reranker Training}

To improve semantic ranking of retrieved PubMed documents, we fine-tuned a SentenceTransformer-based MiniLM reranker using the BioASQ training collection. Positive training examples consisted of BioASQ question–document pairs annotated as relevant, while negative examples were constructed from retrieved documents that were not annotated as relevant for the corresponding question. The resulting dataset was randomly divided into 90\% training and 10\% validation. The reranker was optimized for 10 epochs using MultipleNegativesRankingLoss with Adam optimization and a linear warm-up schedule. During inference, the fine-tuned model computes semantic similarity scores between each question and candidate documents, which are combined with heuristic relevance signals before Reciprocal Rank Fusion.

\subsection{Answer Generation and Post-processing}

For answer generation, we employ OpenBioLLM with type-specific prompting strategies
tailored for yes/no, factoid, list, and summary questions. Distinct prompt templates are designed to elicit concise exact answers for factoid and yes/no questions while permitting more elaborate explanatory responses for summary-type ideal answers. In Phase A+, prompts are conditioned on the documents and snippets retrieved by our retrieval pipeline. Phase B operates on the curated evidence documents and snippets provided by BioASQ. This separation enables the answer generation component to be assessed under both retrieval-dependent and retrieval-independent settings, providing a cleaner ablation of retrieval quality from generation quality.

\subsubsection{Model Selection and Quantization}

We adopt \texttt{aaditya/Llama3-OpenBioLLM-70B} as the generative backbone
for Phase~B. The 70-billion parameter model provides substantially greater reasoning capacity and domain knowledge than its 8B counterpart, which is critical for synthesising accurate biomedical answers from potentially sparse or conflicting snippet evidence.

To deploy a 70B model within the memory envelope of a single NVIDIA A100 40\,GB GPU, we
apply \textbf{Activation-aware Weight Quantization} (AWQ)~\cite{lin2023}. AWQ identifies a small fraction (${\sim}1\%$) of \emph{salient} weight channels, those corresponding to large activation magnitudes and protects them during quantisation via per-channel scaling factors, while compressing the remaining weights to 4-bit precision. This strategy yields substantially better output fidelity than naive round-to-nearest (RTN) quantisation at comparable bit-widths. The resulting memory footprint is as follows:
\begin{itemize}
    \item \textbf{Weight storage:} $70 \times 10^{9} \times 0.5\,\text{bytes} = 35.0\,\text{GB}$
    \item \textbf{Group quantisation metadata} (group size $= 128$, FP16 scales): ${\approx}1.1\,\text{GB}$
    \item \textbf{Total model footprint:} ${\approx}36.1\,\text{GB}$ at 4-bit precision
\end{itemize}
The remaining ${\approx}3.9\,\text{GB}$ of the 40\,GB budget is allocated to the key-value
(KV) cache, CUDA workspace tensors, and activation buffers. On a A100 hardware, AWQ exploits Tensor Core support for INT4 matrix multiplication, achieving near-lossless throughput relative to FP16 inference while reducing memory bandwidth requirements by $4\times$.

\subsubsection{Inference Engine: vLLM with PagedAttention}

The quantized model is served using vLLM~\cite{kwon2023}, a high-throughput inference engine that supports efficient deployment of large language models through PagedAttention and continuous batching. PagedAttention manages the key-value (KV) cache using fixed-size memory blocks, substantially reducing memory fragmentation and improving GPU memory utilization compared with conventional contiguous allocation. Although our pipeline processes questions sequentially (batch size = 1), vLLM still provides low-overhead inference and stable execution.

To ensure reliable deployment on a single NVIDIA A100 40\, GB GPU, the inference engine was configured conservatively using:
\begin{align*}
    \texttt{gpu\_memory\_utilization} &= 0.95 \quad \text{(38\,GB allocated)} \\
    \texttt{max\_model\_len}          &= 2048  \quad \text{(prompt + generation token budget)} \\
    \texttt{enforce\_eager}           &= \texttt{True} \quad \text{(CUDA graphs disabled, saving ${\approx}1$\,GB)} \\
    \texttt{tensor\_parallel\_size}   &= 1     \quad \text{(single-GPU deployment)}
\end{align*}
Setting \texttt{enforce\_eager=True} disables vLLM's CUDA graph optimisation, trading a
small constant-factor throughput reduction for approximately 1\,GB of additional usable
memory.

\subsubsection{Prompt }

Each question is transformed into a structured prompt comprising of four ordered components:
\begin{itemize}
    \item[i] a \emph{question type marker} that signals the expected output format
    \item[ii] the raw \emph{question body}
    \item[iii] up to three \emph{gold snippets} annotated with provenance markers (\texttt{[Snippet 1]}, \texttt{[Snippet 2]}, etc.)
    \item[iv] \emph{type-specific formatting instructions} that enforce BioASQ-compliant output structure. 
\end{itemize}
Type-specific instructions are calibrated to the BioASQ output schema: yes/no questions
require an "\texttt{EXACT: yes}" or "\texttt{EXACT: no}" prefix followed by \texttt{IDEAL:~\ldots}. Factoid and list questions expect pipe-delimited entities
(\texttt{EXACT: entity1 | entity2}). Summary questions require only the \texttt{IDEAL:}
prefix with a 2--3 sentence response. The \texttt{Your answer:} delimiter provides an
unambiguous extraction boundary for downstream post-processing. This structured approach simultaneously constrains the model's output space to the evaluation schema and furnishes sufficient in-context specification to promote format compliance without requiring few-shot exemplars.

\subsubsection{Evaluation Protocol}

System outputs are evaluated against gold-standard answers using type-specific metrics
\cite{tsatsaronis2015}.

\paragraph{Yes/No questions.}
Performance is measured via macro-averaged $F_1$:
\begin{equation}
    F_1^{\text{macro}} = \frac{1}{2}\!\left(F_1^{\text{yes}} + F_1^{\text{no}}\right),
\end{equation}
where $F_1^{c}$ denotes the per-class $F_1$ score obtained by treating class $c$ as the
positive label.

\paragraph{Factoid questions.}
Three complementary metrics are reported:
\begin{itemize}
    \item \textbf{Strict accuracy} - 1 if the top-ranked prediction matches any accepted
          gold synonym, else 0.
    \item \textbf{Lenient accuracy} - 1 if any prediction within the top-5 list matches a
          gold synonym.
    \item \textbf{Mean Reciprocal Rank (MRR)} -
          $\displaystyle\frac{1}{N}\sum_{i=1}^{N}\frac{1}{\mathrm{rank}_i}$,
          where $\mathrm{rank}_i$ is the rank of the first correct prediction for question
          $i$, or 0 if no correct prediction is returned.
\end{itemize}

\paragraph{List questions.}
Evaluation is based on mean precision, recall, and $F_1$ across all questions:
\begin{equation}
    P = \frac{|C|}{|R|}, \qquad R = \frac{|C|}{|T|}, \qquad
    F_1 = \frac{2PR}{P + R},
\end{equation}
where $C$ denotes the set of correct predictions, $R$ the set of returned items, and $T$
the complete set of gold items. Item matching employs synonym-aware comparison, whereby a
prediction is considered correct if it corresponds to any accepted synonym of a gold entity.
\paragraph{Ideal Answer.} Abstractive-answer quality is evaluated using ROUGE-based overlap metrics against the BioASQ reference ideal answers. Following the official BioASQ evaluation protocol, we report ROUGE-2 F1 and ROUGE-SU4 F1 scores. ROUGE-2 measures bigram overlap between generated and reference summaries, while ROUGE-SU4 additionally incorporates skip-bigram matching with a maximum distance of four words, providing a more flexible assessment of semantic and structural similarity in generated biomedical summaries.
\section{Results}
\subsection{MiniLM Reranker Training}

Following the training procedure described in Section 4.3, we evaluated the convergence behavior of the fine-tuned SentenceTransformer-based MiniLM reranker. Figure 3 illustrates the training and validation loss curves over 10 training epochs. The steadily decreasing training loss and stable validation loss demonstrate effective optimization and good generalization of the reranker on BioASQ question–document relevance prediction.

\begin{figure}[htbp]
    \centering
    \includegraphics[width=0.8 \linewidth]{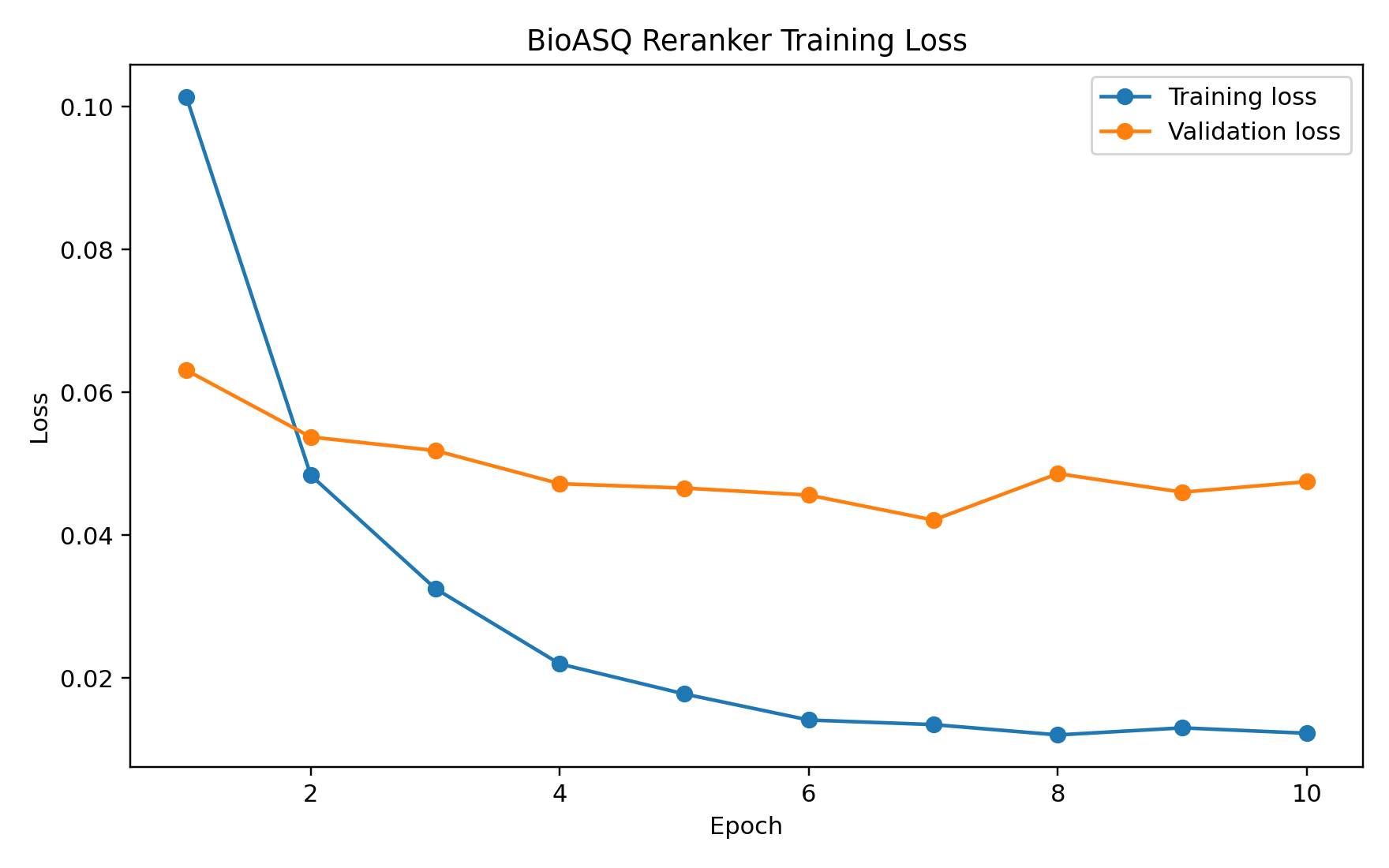}
    \caption{Training and validation loss curves for the fine-tuned MiniLM-based reranker across 10 training epochs. The training loss decreases steadily while the validation loss stabilizes after early epochs, indicating stable convergence with limited overfitting.}
    \label{fig:placeholder}
\end{figure}

As shown in Figure~3, the training loss decreases sharply during the first few epochs, dropping from approximately 0.10 to below 0.05 by epoch 2. This rapid reduction indicates that the model quickly learns general relevance patterns from the training data. After this initial phase, the training loss continues to decline more gradually, reaching a stable level at around 0.01–0.02, suggesting convergence.

The validation loss exhibits a smoother trend, decreasing from approximately 0.063 to around 0.042 by epoch 7. This indicates consistent improvement in generalization during early training. After epoch 7, the validation loss begins to fluctuate slightly, with a small increase observed around epoch 8, suggesting the onset of mild overfitting. However, the gap between training and validation loss remains relatively small throughout training, indicating that overfitting is limited and well controlled.

\FloatBarrier
\subsection{Phase A Result Overview}

Table 1 summarizes the preliminary official BioASQ Phase A retrieval results across the four evaluation batches. Among the submitted runs with available official results, Batch 4 achieved the strongest performance for our system, with Recall = 0.1539, F-measure = 0.0926, and MAP = 0.0956. While the highest-performing submissions achieved MAP values above 0.23 on Batch 4, our retrieval pipeline consistently improved over the baseline through query expansion, MiniLM reranking, weak-question recovery, and retrieval cleanup. Batch 3 exhibited lower retrieval performance, likely reflecting more challenging question formulations with weaker lexical overlap or greater reliance on domain-specific terminology. These observations motivated the introduction of additional weak-question refinement and retrieval recovery strategies, whose effectiveness is demonstrated through the baseline-to-final retrieval pipeline comparison presented in Table 2.

\begin{table}[ht]
\centering
\small
\caption{Phase A retrieval performance across BioASQ Task 14a batches.}
\label{tab:phaseA_results}
\begin{tabular}{lcccc}
\toprule
\textbf{Batch} & \textbf{Recall} & \textbf{Mean Precision} & \textbf{MAP} & \textbf{F-Measure} \\
\midrule
Batch 1$^\ast$ & 0.1702 & 0.0463 & 0.1062 & 0.0727 \\
Batch 2 & 0.1249 & 0.0642 & 0.0917 & 0.0728 \\
Batch 3 & 0.1000 & 0.0200 & 0.0694 & 0.0326 \\
Batch 4 & \textbf{0.1539} & \textbf{0.0733} & \textbf{0.0956} & \textbf{0.0926} \\
\bottomrule
\end{tabular}

\vspace{1mm}
\footnotesize{
$^\ast$ Batch 1 values were recalculated locally using the final pipeline after correcting a submission formatting issue.
}
\end{table}

As shown in Table 2, the local proxy comparison between the baseline and final retrieval pipelines indicates that the complete retrieval framework, including weak-question recovery and retrieval cleanup, achieved a MAP@10 of 0.0955 compared with 0.0745 for the corresponding baseline retrieval pipeline, representing a relative improvement of approximately 28\%. The final pipeline also increased Recall@10 from 0.1335 to 0.1539, suggesting that the retrieval refinements enabled the system to recover relevant biomedical documents that were missed by the baseline strategy. Although this comparison was performed using our local evaluation script together with the released Phase B evidence, the resulting performance closely matches the official Batch 4 BioASQ evaluation (MAP = 0.0956), providing additional confidence that the proposed retrieval refinements generalize beyond the local evaluation setting. Because the retrieval refinements were evaluated as an integrated pipeline, the reported improvements reflect the combined contribution of query expansion, neural reranking, weak-question recovery, retrieval fusion, and retrieval cleanup rather than the isolated effect of any individual component.
\begin{table}[ht]
\centering
\small
\caption{Local proxy ablation of Batch 4 weak-question recovery}
\label{tab:phaseA_weak_rescue_ablation}
\begin{tabular}{lcccc}
\hline
\textbf{Configuration} & \textbf{Recall@10} & \textbf{Precision@10} & \textbf{MAP@10} & \textbf{F1} \\
\hline
Batch 4 Base & 0.1335 & 0.0383 & 0.0745 & 0.0595 \\
Batch 4 Final + Rescue & 0.1539 & 0.0367 & 0.0955 & 0.0594 \\
\hline
\end{tabular}
\end{table}

To examine whether larger reranking models could improve retrieval performance, we evaluated several alternative reranking configurations on Batch 4. These included a larger pretrained MPNet bi-encoder, a BioASQ fine-tuned MPNet model, and the publicly available pretrained MS MARCO MiniLM cross-encoder. As shown in Table 3, these alternatives did not outperform the final MiniLM-based system. The final system achieved the highest Recall@10, Precision@10, and MAP@10 among the evaluated reranking configurations, suggesting that the lightweight MiniLM reranker, when combined with the complete retrieval pipeline, provided the best trade-off between retrieval effectiveness and computational efficiency. These results indicate that simply increasing model size or adopting a generic pretrained cross-encoder was insufficient to improve BioASQ retrieval performance. Instead, combining domain-specific reranker fine-tuning with query expansion, retrieval fusion, weak-question recovery, and retrieval cleanup produced the strongest overall retrieval performance.

\begin{table}[ht]
\centering
\caption{Comparison of reranking and refinement variants on the Batch 4 evaluation split.}
\label{tab:reranker_ablation}
\begin{tabular}{lccc}
\hline
\textbf{Configuration} & \textbf{Recall@10} & \textbf{Precision@10} & \textbf{MAP@10} \\
\hline
Final MiniLM-based system & 0.1539 & 0.0733 & 0.0956 \\
MPNet bi-encoder reranker & 0.1454 & 0.0400 & 0.0836 \\
Fine-tuned MPNet reranker & 0.1377 & 0.0400 & 0.0820 \\
MS MARCO MiniLM cross-encoder & 0.1281 & 0.0400 & 0.0675 \\
\hline
\end{tabular}
\end{table}

Although our retrieval performance did not reach the highest-performing BioASQ submissions, the proposed retrieval pipeline consistently improved over the baseline while maintaining a lightweight and computationally efficient architecture. The close agreement between the local proxy evaluation (MAP = 0.0955) and the official Batch 4 BioASQ result (MAP = 0.0956) further supports the reliability of the proposed evaluation methodology for guiding retrieval system development.

\subsection{Phase A+ Result Overview}
Preliminary official Phase A+ results are summarized in Table 4. Exact-answer performance improved across the later evaluation batches. Batch 4 achieved the strongest overall performance for our system, reaching a yes/no macro-F1 of 0.8667, a list-question F1 of 0.2588, and factoid strict, lenient, and MRR scores of 0.1818, 0.2727, and 0.2273, respectively. Factoid performance peaked in Batch 3, where strict accuracy, lenient accuracy, and MRR all reached 0.3529, indicating that the proposed retrieval and answer-normalization pipeline was particularly effective for entity-oriented biomedical questions when relevant evidence was successfully retrieved.

Among all participating systems, our submission achieved mid-table rankings across the four official batches. While the best-performing systems generally obtained higher factoid and list-answer scores through more advanced retrieval or answer-generation strategies, our lightweight retrieval-augmented pipeline consistently produced competitive exact-answer performance while maintaining a relatively simple architecture based on query expansion, MiniLM reranking, weak-question recovery, and OpenBioLLM generation. The improvements observed from Batch 1 through Batch 4 also demonstrate that the retrieval refinements introduced during system development translated into measurable gains on the official evaluation batches.
\begin{table*}[ht]
\centering
\small
\caption{Preliminary official Phase A+ results for DS@GT-BioASQ.}
\label{tab:phaseAplus_combined}
\resizebox{\textwidth}{!}{
\begin{tabular}{lccccccccc}
\hline
\multirow{2}{*}{\textbf{Batch}} &
\multicolumn{7}{c}{\textbf{Exact Answers}} &
\multicolumn{2}{c}{\textbf{Ideal Answers}} \\
\cline{2-10}

& \textbf{YN Macro F1}
& \textbf{Factoid Strict}
& \textbf{Factoid Lenient}
& \textbf{Factoid MRR}
& \textbf{List Prec.}
& \textbf{List Recall}
& \textbf{List F1}
& \textbf{R-2 F1}
& \textbf{R-SU4 F1} \\

\hline

Batch 1 & 0.6458 & 0.0435 & 0.0435 & 0.0435 & 0.2024 & 0.1148 & 0.1402 & 0.1183 & 0.1284 \\
Batch 2 & 0.4000 & 0.1500 & 0.1500 & 0.1500 & 0.2674 & 0.1442 & 0.1707 & \textbf{0.1583} & \textbf{0.1540} \\
Batch 3 & 0.4211 & \textbf{0.3529} & \textbf{0.3529} & \textbf{0.3529} & 0.2923 & 0.1787 & 0.2021 & 0.1151 & 0.1174 \\
Batch 4 & \textbf{0.8667} & 0.1818 & 0.2727 & 0.2273 & \textbf{0.3592} & \textbf{0.2249} & \textbf{0.2588} & 0.1145 & 0.1152 \\

\hline
\end{tabular}
}
\end{table*}

Ideal-answer ROUGE scores were comparatively more stable across batches but remained lower than the exact-answer metrics. The highest ideal-answer performance was observed in Batch 2, achieving ROUGE-2 F1 = 0.1583 and ROUGE-SU4 F1 = 0.1540. Compared with the top-performing systems, which achieved ROUGE-2 F1 values of approximately 0.15–0.16, our system produced lower abstractive-answer quality despite competitive retrieval performance. This suggests that retrieval refinements contributed more substantially to exact-answer prediction than to abstractive summary generation. Future improvements will therefore focus on stronger biomedical instruction tuning, retrieval-aware prompting, and citation-aware evidence synthesis to improve long-form answer generation while preserving factual grounding.

Overall, the Phase A+ results demonstrate that the proposed hybrid retrieval and OpenBioLLM-based generation framework produces structurally valid and evidence-grounded biomedical answers across multiple BioASQ evaluation batches, while highlighting remaining opportunities for improving abstractive answer generation.

\subsection{Phase B Result Overview}
Preliminary official Phase B results are summarized in Table 5. Overall, exact-answer performance improved compared with Phase A+, particularly for yes/no and factoid questions, reflecting the advantage of using the curated BioASQ evidence documents provided in the Phase B setting. Batch 3 achieved the strongest overall exact-answer performance for our system, with a yes/no macro-F1 of 0.8952 and factoid strict accuracy, lenient accuracy, and MRR all reaching 0.4118. List-question performance also remained stable across later batches, with Batch 4 achieving the highest list precision and F1 scores of 0.4125 and 0.3020, respectively.

Compared with the top-performing Phase B submissions, which achieved near-perfect yes/no performance and stronger factoid and list-answer scores, our system showed consistent improvements over the corresponding Phase A+ setting while maintaining competitive mid-table performance on the official Phase B leaderboard. These results demonstrate that the proposed retrieval-augmented OpenBioLLM framework generalizes across multiple biomedical question types despite its relatively lightweight architecture. The improvements observed in Phase B further suggest that the OpenBioLLM-based answer generation and post-processing pipeline can effectively leverage high-quality biomedical evidence when relevant documents and snippets are provided.
\begin{table*}[ht]
\centering
\small
\caption{Preliminary official Phase B results for DS@GT-BioASQ.}
\label{tab:phaseB_combined}
\resizebox{\textwidth}{!}{
\begin{tabular}{lccccccccc}
\hline
\multirow{2}{*}{\textbf{Batch}} &
\multicolumn{7}{c}{\textbf{Exact Answers}} &
\multicolumn{2}{c}{\textbf{Ideal Answers}} \\
\cline{2-10}
& \textbf{YN Macro F1}
& \textbf{Factoid Strict}
& \textbf{Factoid Lenient}
& \textbf{Factoid MRR}
& \textbf{List Prec.}
& \textbf{List Recall}
& \textbf{List F1}
& \textbf{R-2 F1}
& \textbf{R-SU4 F1} \\
\hline
Batch 1 & 0.7571 & 0.2609 & 0.2609 & 0.2609 & 0.3119 & 0.2010 & 0.2285 & 0.1421 & 0.1485 \\
Batch 2 & 0.8329 & 0.2000 & 0.2000 & 0.2000 & 0.2888 & 0.2730 & 0.2783 & 0.1252 & 0.1233 \\
Batch 3 & \textbf{0.8952} & \textbf{0.4118} & \textbf{0.4118} & \textbf{0.4118} & 0.3407 & \textbf{0.2659} & 0.2939 & \textbf{0.1758} & \textbf{0.1655} \\
Batch 4 & 0.8667 & 0.1818 & 0.2727 & 0.2273 & \textbf{0.4125} & 0.2565 & \textbf{0.3020} & 0.1645 & 0.1475 \\
\hline
\end{tabular}
}
\end{table*}

Ideal-answer ROUGE performance was more stable across batches. Batch 3 achieved the highest ROUGE-2 F1 and ROUGE-SU4 F1 scores, reaching 0.1758 and 0.1655, respectively, indicating improved alignment between generated summaries and the BioASQ reference ideal answers. Compared with Phase A+, the Phase B results suggest that access to curated gold evidence substantially improved exact-answer prediction while also providing moderate gains for abstractive answer generation.

Overall, the Phase B results indicate that evidence quality is a major determinant of downstream answer generation performance. When high-quality supporting evidence is available, the proposed OpenBioLLM-based generation framework produces more reliable exact answers and moderately stronger ideal answers, highlighting the importance of improving retrieval quality in the full end-to-end Phase A+ setting.

\section{Discussion}

One of the clearest observations during the BioASQ submission rounds was that retrieval failures were often concentrated within a relatively small subset of difficult biomedical questions. These weak-performing queries frequently involved rare biomedical terminology, limited lexical overlap with relevant PubMed articles, or complex multi-hop relationships between biomedical entities. Rather than globally modifying the entire retrieval pipeline, we introduced a conditional weak-question recovery strategy that selectively revisited underperforming queries using additional refinement and retrieval cleanup steps.

The Batch~4 experiments suggest that this targeted recovery strategy improved retrieval robustness for difficult biomedical questions. In particular, the refinement pipeline increased both Recall@10 and MAP@10 relative to the corresponding base retrieval configuration, indicating that the system was able to recover relevant biomedical evidence that was previously missed while also improving the ranking quality of retrieved documents. The larger improvement observed in MAP compared to F1 further suggests that the refinement process primarily improved the ordering of relevant evidence near the top of the ranked list, which is particularly important for downstream biomedical question answering tasks.

More broadly, these observations highlight the importance of adaptive retrieval refinement in biomedical QA systems. Difficult biomedical questions often contain specialized terminology, implicit entity relationships, or limited lexical overlap with relevant literature, making them challenging for standard retrieval pipelines. The experiments suggest that selectively targeting weak or underperforming queries can produce meaningful retrieval improvements without requiring substantial modifications to the core retrieval architecture.

The results also indicate that lightweight reranking approaches combined with targeted retrieval refinement may provide a stronger tradeoff between effectiveness and computational efficiency than simply increasing model size. In our experiments, larger reranking models and generic cross-encoder architectures did not consistently outperform the final MiniLM-based retrieval pipeline when weak-question recovery and evidence filtering were incorporated.

Overall, the proposed framework demonstrates that retrieval-aware refinement and controlled evidence selection can improve retrieval stability and evidence grounding for biomedical question answering tasks, particularly for difficult or low-overlap biomedical queries.

As for answer generation in Phase B, one important observation from this work is that retrieval quality alone does not necessarily guarantee correct biomedical answers. The results indicate that answer quality was highly dependent on the quality of the retrieved evidence. This observation is consistent with the Phase B results, where providing curated BioASQ evidence substantially improved answer quality without modifying the generation model itself. Although the retrieval pipeline generally returned topically relevant biomedical documents, successful answer generation often required evidence that explicitly contained the target entity, relationship, or clinical concept. This challenge was particularly apparent for factoid and list questions, where omission of a single biomedical entity could substantially affect strict accuracy, MRR, or list F1 scores.

Controlled evidence selection also played an important role in downstream answer generation quality. Excessive snippet retention often introduced redundant or weakly relevant biomedical evidence, while overly aggressive pruning reduced evidence coverage and negatively affected recall-oriented questions. The experiments suggest that maintaining a limited set of high-confidence evidence snippets provided a better balance between evidence coverage and generation stability for biomedical QA.

Another important observation is that formatting robustness remains critical for biomedical benchmark evaluations such as BioASQ. A substantial portion of early submission failures originated from malformed exact-answer structures, missing fields, or inconsistent formatting rather than incorrect biomedical reasoning. The addition of post-processing and answer normalization substantially improved submission validity and evaluation stability across batches.

The official preliminary Phase B results demonstrate relatively stable performance across multiple biomedical question types. Yes/no questions achieved the strongest overall performance, with macro F1 scores ranging from 0.7571 to 0.8952 across batches. Factoid performance varied more substantially across batches and appeared particularly sensitive to entity coverage within the retrieved evidence, with the strongest Batch 3 performance reaching 0.4118 for strict accuracy, lenient accuracy, and MRR. List-question performance remained comparatively stable, with Batch~4 achieving the strongest list F1 score of 0.3020. For ideal-answer generation, ROUGE-2 F1 and ROUGE-SU4 F1 scores remained relatively consistent across batches, suggesting that the OpenBioLLM-based generation pipeline produced stable abstractive biomedical summaries despite variations in retrieval difficulty.

Overall, these observations highlight the importance of combining adaptive retrieval refinement, controlled evidence selection, and robust answer post-processing in biomedical question answering systems. The results suggest that targeted improvements for difficult biomedical queries can improve both retrieval stability and downstream answer quality without requiring major modifications to the core retrieval architecture.
\section{Future Work}

Several directions could further improve the proposed system. First, incorporating few-shot or retrieval-aware prompting strategies may improve answer consistency and grounding quality across different biomedical question types. Although the current answer generation system uses type-specific prompting templates, the prompts remain relatively static and may benefit from adaptive prompt selection based on retrieval confidence or evidence diversity.

Future work could also explore stronger biomedical reranking models trained directly on BioASQ-style supervision. While the current MiniLM-based reranker provides efficient semantic matching, domain-specific reranking trained on BioASQ relevance annotations or hard-negative mining may further improve retrieval precision for challenging biomedical entities and rare terminology. In addition, although the current retrieval refinements were evaluated as an integrated pipeline, future work will perform comprehensive component-wise ablation studies to quantify the individual contributions of query expansion, MiniLM reranking, weak-question recovery, and retrieval cleanup. Such analysis will provide a clearer understanding of how each retrieval component contributes to retrieval effectiveness and downstream biomedical question answering performance.

Another promising direction is improving alignment between retrieved evidence and generated answers. While the current proxy grounding analysis demonstrates reasonable evidence overlap, hallucination and unsupported reasoning remain important challenges for biomedical LLMs. Integrating confidence estimation, evidence attribution, or citation-aware generation may improve interpretability and reliability for downstream biomedical applications.

Finally, future systems could explore more advanced agentic retrieval strategies and ontology-guided retrieval augmentation. Future work will also investigate integrating larger biomedical foundation models together with adaptive retrieval planning to further improve multi-hop biomedical reasoning. Recent biomedical RAG frameworks suggest that iterative retrieval refinement and structured biomedical knowledge integration may further improve robustness for complex multi-hop biomedical reasoning tasks. Preliminary experiments with dynamic knowledge-graph–assisted retrieval did not produce measurable improvements under the current evaluation setting; however, more comprehensive graph construction and relation-aware reasoning remain promising directions for future investigation.

\section{Conclusions}
This work presents an end-to-end biomedical question answering pipeline for BioASQ Task 14b that integrates retrieval, reranking, OpenBioLLM-based answer generation, and post-processing into a unified framework. The proposed system combines semantic query expansion, weak-question recovery, retrieval cleanup, and structured output validation to improve both retrieval quality and submission robustness.

The results demonstrate that effective biomedical question answering depends not only on strong retrieval and generation models, but also on careful system-level engineering decisions. In particular, selective retrieval refinement, controlled snippet pruning, and structured post-processing substantially improved evidence grounding and formatting stability across BioASQ evaluation batches.

Overall, this work highlights the importance of combining retrieval-aware reasoning, evidence-focused generation, and robust output validation for biomedical QA systems. The proposed framework demonstrates that lightweight retrieval refinement, neural reranking, and evidence-aware answer generation can produce reliable biomedical question answering while remaining computationally efficient.

\section*{Acknowledgements}

We thank the Data Science at Georgia Tech (DS@GT) CLEF competition group for their support.
This research was supported in part through research cyberinfrastructure resources and services provided by the Partnership for an Advanced Computing Environment (PACE) at the Georgia Institute of Technology, Atlanta, Georgia, USA \cite{PACE}. 

%% The declaration on generative AI comes in effect
%% in Janary 2025. See also
%% https://ceur-ws.org/GenAI/Policy.html
\section*{Declaration on Generative AI}
During the preparation of this work, the authors used ChatGPT (OpenAI GPT-based models) for language refinement, grammar checking, and manuscript editing support. The authors reviewed, revised, and validated all generated content and take full responsibility for the final manuscript.
\newline

% Need to site BioASQ and OpenBioLLM

\bibliography{main}
\end{document}